\documentclass[letterpaper, 10 pt, journal, twoside]{ieeetran}  %

\usepackage{amsmath,amssymb,amsfonts}
\usepackage{algorithmic}
\usepackage{graphicx}
\usepackage{textcomp}
\usepackage{xcolor}
\usepackage{makecell}
\usepackage{multirow} 
\usepackage{tabularx} 
\usepackage{caption}
\usepackage{siunitx}
\usepackage{enumitem}
\usepackage{hhline}
\usepackage{siunitx}
\usepackage{booktabs} 
\usepackage{subcaption}
\usepackage{hyperref}
\usepackage{rotating}

\title{EP-Diffuser: An Efficient Diffusion Model for Traffic Scene Generation and Prediction via Polynomial Representations
}

\author{Yue Yao$^{1}$, Mohamed-Khalil Bouzidi$^{1}$, Daniel Goehring$^1$, Joerg Reichardt$^2$
\thanks{Manuscript received: March 03, 2025; Revised: May 21, 2025; Accepted: July 08, 2025.} 
\thanks{This paper was recommended for publication by Editor Lucia Pallottino upon evaluation of the Associate Editor and Reviewers' comments.} 
\thanks{$^{1}$Yue Yao, Mohamed-Khalil Bouzidi and Daniel Goehring are with the Department of Mathematics and Computer Science, Freie Universität Berlin, Germany
{\tt\footnotesize (yue.yao@fu-berlin.de; mohamed-khalil.bouzidi@fu-berlin.de; daniel.goehring@fu-berlin.de)}}
\thanks{$^{2}$Joerg Reichardt is with Continental Automotive GmbH, Germany {\tt\footnotesize(joerg.reichardt@continental.com)}}
\thanks{Digital Object Identifier (DOI): see top of this page.}
}

\begin{document}

\markboth{IEEE Robotics and Automation Letters. Preprint Version. Accepted July, 2025}
{Yao \MakeLowercase{\textit{et al.}}: Efficient Traffic Scene Generation via Polynomial Representations} 

\maketitle

\begin{abstract}
As the prediction horizon increases, predicting the future evolution of traffic scenes becomes increasingly difficult due to the multi-modal nature of agent motion. Most state-of-the-art (SotA) prediction models primarily focus on forecasting the most likely future. However, for the safe operation of autonomous vehicles, it is equally important to cover the distribution for plausible motion alternatives. To address this, we introduce EP-Diffuser, a novel parameter-efficient diffusion-based generative model designed to capture the distribution of possible traffic scene evolutions. Conditioned on road layout and agent history, our model acts as a predictor and generates diverse, plausible scene continuations. We benchmark EP-Diffuser against two SotA models in terms of plausibility, diversity, and accuracy of predictions on the Argoverse 2 dataset. Despite its significantly smaller model size, our approach achieves both highly plausible and diverse traffic scene predictions with comparable accuracy. We further evaluate model generalization in an out-of-distribution (OoD) test setting using Waymo Open dataset and show superior robustness of our approach.
The code and model checkpoints are available at: \href{https://github.com/continental/EP-Diffuser}{https://github.com/continental/EP-Diffuser}.
\end{abstract}


\section{Introduction} 
\IEEEPARstart{T}{raffic} is a complex phenomenon where multiple agents interact in shared space and influence each other's behavior. For autonomous vehicles to integrate safely into such dynamic environments, they must anticipate how traffic scenes will evolve over time. This prediction capability is essential for downstream planning and decision-making processes.

Public motion datasets, such as Argoverse 2 (A2) \cite{wilson_argoverse2_2021} and Waymo Open (WO) \cite{ettinger_waymo_2021}, provide real-world traffic scene data and host associated motion prediction competitions to advance research in this field. The evolution of traffic scenes over long time horizons is governed by an inherently multi-modal probability distribution. Multiple plausible futures exist depending on road topology and agent interactions. However, motion datasets can only record a single observed future sample (the ground truth) per scene. As a result, motion prediction competitions typically frame the problem as a \emph{regression} problem, where models are trained to estimate the most likely outcome based on available ground truth data. Existing approaches can be categorized into two main types:
\begin{itemize}[leftmargin=*]
   \item \emph{Marginal Prediction}: Forecasting individual agent trajectories without ensuring that they combine into consistent scenes.
   \item \emph{Joint Prediction}: Modeling
multiple interacting agents simultaneously to predict coherent traffic scene continuations, a task we refer to as \emph{traffic scene prediction}.
\end{itemize}

\begin{figure}[t]
\centering
\includegraphics[width=2.8in]{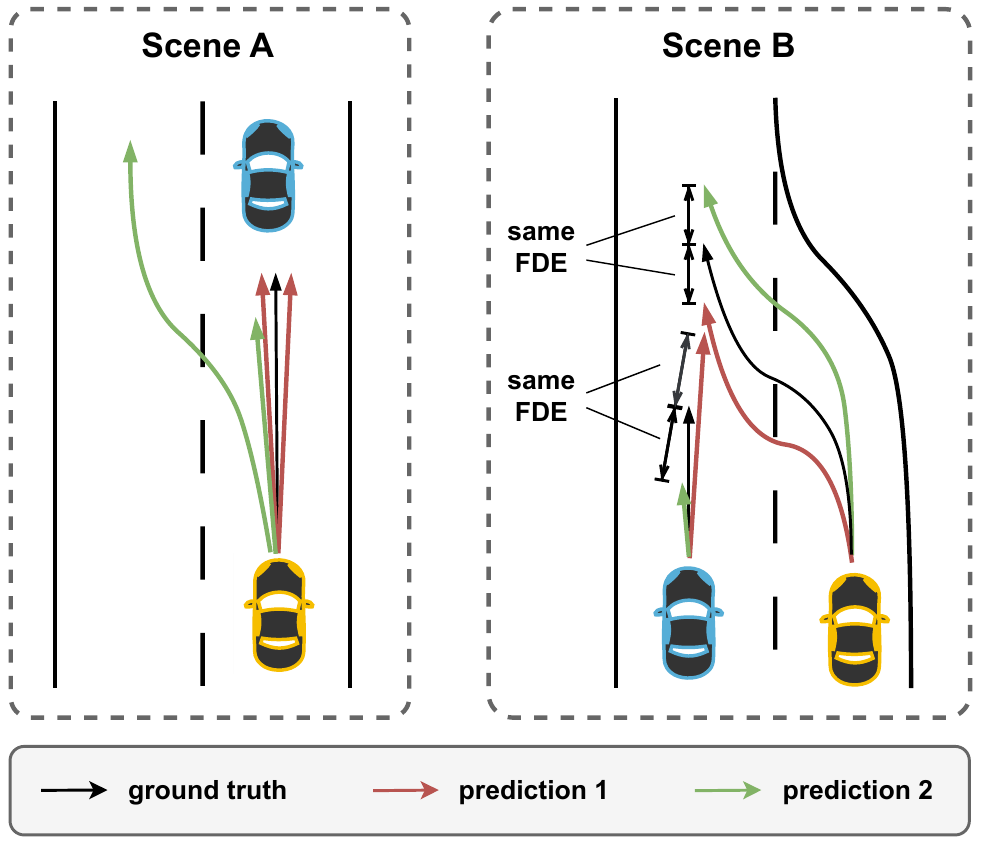}
\caption{Two limitations of regression-based metrics. \textbf{Scene A}: An example of \emph{multi-modal marginal} prediction. While Prediction 1 yields a lower Final Displacement Error (FDE), it only captures one possible behavior and ignores other plausible maneuvers. \textbf{Scene B}: An example of \emph{uni-modal joint} prediction. While both predictions yield the same FDE, Prediction 1 is less plausible due to the agent collision.}
\label{fig: regression metrics issue}
\vspace{-1.5em}
\end{figure}

\noindent Regression-based metrics, such as Average Displacement Error (ADE), Final Displacement Error (FDE), and their variants, are widely used in these competitions. These metrics evaluate prediction accuracy by measuring how closely predictions match ground truth trajectories. However, this evaluation approach presents two key limitations illustrated in Figure \ref{fig: regression metrics issue}: First, it does not measure the diversity and coverage of predictions. Second, it fails to account for plausibility and consistency, both of which are essential for safe trajectory planning. Furthermore, focusing solely on the most probable future evolution while ignoring other plausible possibilities may induce overconfident and risky behavior, making it insufficient for planning algorithms, as highlighted in recent trajectory planning studies \cite{chen_interactive_2022, bouzidi_motion_2024, mustafa_racp_2024, bouzidi_closing_2025}.

Regression-based approaches have attempted to improve prediction diversity using discrete trajectory sets \cite{phan_covernet_2020} or modified training objectives \cite{xu_annealed_2024, lidard_nashformer_2023}. In parallel, other studies have explored \emph{generative} modeling frameworks. These include methods based on GANs \cite{gupta_social_2018, roy_vehicle_2019}, VAEs \cite{salzmann_trajectron_2020}, and diversity-enhancing sampling methods \cite{huang_diversitygan_2020, ma_diverse_2020}, which initially aim to produce diverse marginal predictions for individual agents. 

Building on this trend, recent generative approaches have begun addressing the more challenging task of multi-agent \emph{traffic scene generation}. These models aim to learn the distribution of future traffic scenes under given conditions (i.e., observed agent history and road layout) and enable sampling from the modeled joint distribution. Therefore, they can capture not only the most likely traffic scene evolution but also a range of plausible alternatives.

However, evaluating generative models is non-trivial, as it requires assessing the modeled distribution rather than the matching to a single ground truth trajectory. Many recent generative studies choose to inherit evaluation metrics from regression-based approaches, despite fundamental differences in modeling objectives \cite{choi_dice_2024, wang_optimizing_2025, jiang_motiondiffuser_2023}. This may bias models towards a narrow range of plausible future evolutions.

Another fundamental challenge is assessing model generalization. Prior studies typically evaluate model performance using a test split from the same dataset used for training. While motion datasets attempt to ensure disjoint splits between training and testing, these subsets still share underlying biases, such as recurring road layouts, traffic flow patterns, and artifacts introduced during data collection and pre-processing. Generative models are known to exhibit memorization -- the ability to produce near-replicas of training data \cite{brown_language_2020, somepalli_understanding_2023}. Consequently, models tested solely on these similar samples may appear to perform well by leveraging their memorizing capability instead of learning robust, transferable representations of traffic patterns. 

Hence, it is essential to rigorously assess generalization ability also for generative models under out-of-distribution (OoD) conditions, where the model cannot rely on memorized examples. Prior studies highlight notable distribution shifts across real-world motion datasets, such as variations in road layouts, traffic densities, and agent behaviors \cite{yao_improving_2024, feng_unitraj_2025}. These shifts present an opportunity to test whether generative models can extrapolate to unseen traffic scenes rather than merely recalling training patterns.

With these considerations in mind, we present Everything Polynomial Diffuser (EP-Diffuser), a generative model for traffic scene generation conditioned on road layout and observed agent history, thereby performing joint prediction. Unlike traditional models that incorporate sequence-based data -- such as lists of trajectory observations or map points -- or those that use polynomials solely as outputs \cite{huang_uncertainty_2019}, our model employs polynomial representations for both map elements and trajectories on the model's input and output sides. To comprehensively evaluate our approach, we extend beyond regression-based metrics and benchmark EP-Diffuser against two state-of-the-art (SotA) models from the perspectives of plausibility, coverage, and accuracy. 
Specifically, we incorporate Waymo’s ``Sim Agents'' metrics \cite{montali_waymo_2024} for evaluating plausibility and assess performance under OoD scenarios from the WO dataset.
Our results demonstrate that the polynomial representation enhances the efficiency of the denoising process, temporal consistency in generated agent kinematics, and generalization in OoD test settings. Our contributions are summarized as follows:
\begin{itemize}[leftmargin=*]
   \item We propose a novel diffusion model for generating diverse and highly realistic traffic scenes on the Argoverse 2 Motion dataset by using polynomial representations.
   \item We compare our model with two SotA models, highlighting a significant disconnect between regression-based metrics and the plausibility of predicted traffic scenes.
   \item We demonstrate superior generalization capabilities of our approach under out-of-distribution (OoD) conditions.
\end{itemize}

\noindent Our paper is organized as follows: We first review recent traffic scene prediction and generation models, along with their evaluation metrics. We then introduce two benchmark models and the “Sim Agents” metrics, highlighting key differences from regression-based metrics. Next, we present our diffusion-based approach with constrained parametric representations. We evaluate our model against benchmarks, analyzing plausibility, coverage, and regression-based performance. Finally, we extend our evaluation to OoD scenes to assess generalization beyond the training data.

\section{Related Work and Preliminaries} 
\subsection{Traffic Scene Prediction and Benchmark Models}

Benchmark datasets and associated prediction competitions have significantly shaped research in traffic scene prediction by framing it as a regression task, evaluating the most likely predicted traffic scenes. Recent studies have followed this competition framework and implemented regression-based deep learning models for traffic scene prediction  \cite{cheng_forecast_2023, zhou_qcnext_2023, luo_jfp_2023}. Although these models can output multiple modes for potential traffic scenes, they are primarily scored and ranked using regression-based metrics such as minimum ADE (minADE) and minimum FDE (minFDE). These variants of ADE and FDE are tailored to multi-modal predictors and measure the minimum displacement error among all predicted modes.

Regression-based approaches have significantly influenced the development of generative models in the field \cite{seff_motionlm_2023, jiang_motiondiffuser_2023, wang_optimizing_2025}. Notably, many diffusion-based models have incorporated regression model backbones to output initial predictions that closely align with ground truth data \cite{mao_leapfrog_2023, wang_optimizing_2025}. Although this design choice effectively optimizes for competition results, it may not fully capture the inherent uncertainty of real-world traffic.

Cross-dataset testing introduces a notable distribution shift \cite{feng_unitraj_2025, yao_improving_2024} and is well suited for evaluating OoD generalization. We focus on models trained on the smaller A2 dataset, enabling a more rigorous assessment of their robustness when applied to scenes from the larger WO dataset. While many open-source models exist for the multi-modal marginal prediction task, there are far fewer open-source benchmark models specifically addressing multi-modal traffic scene (joint) prediction with documented performance and reproducible results. This limitation narrows the pool of suitable benchmark models for our experiment.

\begin{table}[tbh]
\caption{Summary of models under study}
\centering
    \begin{tabularx}{\columnwidth}{c| *{3}{>{\centering\arraybackslash}X} }
    \toprule
    model & FMAE-MA \cite{cheng_forecast_2023}& OptTrajDiff \cite{wang_optimizing_2025} & EP-Diffuser (ours)\\
    \midrule
    input \& output & \multirow{2}{*}{sequence} & \multirow{2}{*}{sequence} & \multirow{2}{*}{polynomial} \\
    representation & && \\
     \midrule
    model type & regression & diffusion & diffusion \\
    \midrule
    \# output samples & 6& inf & inf \\
    \midrule
    \# model parameters &   \multirow{2}{*}{1.9} & \multirow{2}{*}{12.5} & \multirow{2}{*}{3.0} \\
    $[$million$]$ & &  & \\
    \bottomrule
    \end{tabularx}
    \label{tab: model summary}
    \vspace{-10pt}
\end{table}

\begin{figure*}[tbh]
\vspace{+2pt}
\centering
\includegraphics[width=6.2in]{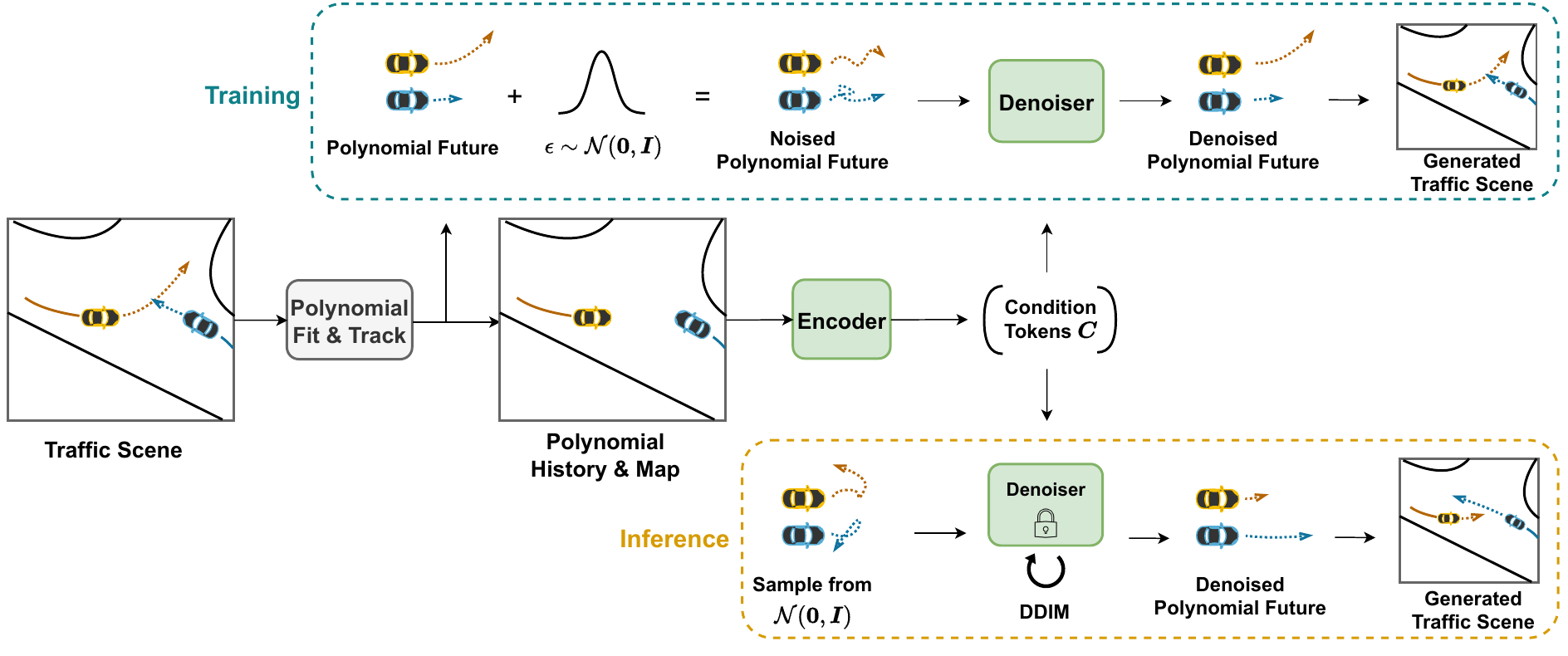}
\caption{Overview of EP-Diffuser for traffic scene generation. For clear presentation, we adopt a similar visual layout to \cite{jiang_motiondiffuser_2023}, while the technical implementation extends our previous EP model \cite{yao_improving_2024}. The traffic scene comprised of agent history and map elements as degree $5$ and $3$ polynomials, respectively, is encoded via a transformer encoder \cite{vaswani_attention_2017} into a set of condition tokens $\boldsymbol{C}$. The ground truth (GT) future trajectories are represented as polynomials of degree $6$. During \textbf{training}, a random set of noise is sampled i.i.d. from a standard normal distribution and added to the parameters of the ground truth future trajectory. The denoiser, while attending to the condition tokens, jointly predicts the denoised polynomial parameters of trajectories corresponding to each agent. During \textbf{inference}, a set of trajectory parameters for each agent is initially sampled from a standard normal distribution, and iteratively denoised using a DDIM schedule\cite{song_denoising_2021} to produce plausible future trajectories.}
\label{fig: diffusion  pipeline}
\vspace{-1.0em}
\end{figure*}

As representatives of the two model classes, we select two recently open-sourced and thoroughly documented SotA models as benchmarks: Forecast-MAE-multiagent (FMAE-MA) \cite{cheng_forecast_2023} and OptTrajDiff \cite{wang_optimizing_2025}. As summarized in Table \ref{tab: model summary}, both models use sequence-based representations but follow different methodological approaches. FMAE-MA follows a regression-based approach and predicts 6 distinct modes of future traffic scenes, with a relatively lightweight architecture of $1.9$ million parameters. OptTrajDiff adopts a diffusion-based approach and incorporates QCNet \cite{zhou_query_2023} as its regression backbone,
with a total of $12.5$ million parameters.

\subsection{Sim Agents Metrics}
\label{sec: sim agents metrics}
In contrast to regression-based tasks, ``Sim Agents'' frames traffic scene prediction as a multi-agent generative task, emphasizing the importance of capturing the diversity and realism of traffic behaviors \cite{montali_waymo_2024}. Rather than focusing solely on minimizing prediction errors, Sim Agents metrics evaluate models based on their ability to produce plausible and consistent traffic scenes using several complementary metrics:
\begin{itemize}[leftmargin=*]
    \item \textbf{Agent Kinematic Metrics}: Evaluate the kinematic properties of individual agents, such as speed, acceleration, and adherence to realistic motion patterns.

    \item \textbf{Agent Interaction Metrics}: Measure the quality of interactions between agents, ensuring that predicted behaviors reflect realistic social dynamics and comply with traffic rules.

    \item \textbf{Map Adherence Metrics}: Assess whether predicted trajectories conform to road layouts, lane boundaries, and other map-related constraints.

    \item \textbf{Realism Meta Metric}: An aggregated score that combines all above evaluations into a holistic measure of scene realism and consistency.
\end{itemize}

\noindent These metrics are calculated by comparing the distribution approximated from 32 predicted samples against real-world data, encouraging models to replicate the variability and interaction patterns observed in actual traffic scenes. All results are normalized scores in the interval $[0,1]$, with 1 indicating the highest score.

In this work, we use the ``realism meta'' metric as the primary measure of predicted scene plausibility.

\section{Data Representation and Model}
EP-Diffuser extends our previous marginal prediction model, EP \cite{yao_improving_2024}, by introducing a generative diffusion framework and enabling multi-agent joint trajectory generation. Unlike other diffusion models that use sequence-based representations \cite{wang_optimizing_2025, mao_leapfrog_2023, choi_dice_2024}, our approach integrates polynomial representations for map elements and trajectories. 

This polynomial design offers three key benefits. First, it provides a low-dimensional, truncated Taylor-like approximation that captures key motion patterns without modeling complex high-order residuals. This improves the efficiency of the diffusion-denoising process and simplifies the overall learning task. Second, since polynomial basis functions possess well-defined derivatives, training with positional loss implicitly influences kinematic terms such as velocity and acceleration. This yields smooth and physically plausible trajectories when appropriately regularized with moderate polynomial degrees. Third, it enhances model generalization by mitigating dataset-specific bias and noise commonly present in sequence-based representations, as highlighted in \cite{yao_improving_2024}. Unlike data-dependent dimensionality reduction methods such as Principal Component Analysis (PCA) used in \cite{jiang_motiondiffuser_2023}, polynomial basis functions are data-independent and support better generalization across datasets.

In the subsequent sections, we first describe our approach for representing diverse data types using polynomial representations, followed by the implementation details of our model.

\subsection{Data as Polynomials}
Different types of polynomials of the same degree -- such as monomial, Bernstein, and Chebyshev polynomials -- can be linearly transformed into one another \cite{reichardt_trajectories_2022}, offering the same expressiveness without affecting the model’s capacity. In this work, we employ Bernstein polynomials to represent agent histories, future trajectories, and map geometry, though our approach is not restricted to this basis. The parameters of Bernstein polynomials have spatial semantics as \emph{control points}, making them particularly convenient for coordinate transformations such as translation, rotation, and scaling.

The A2 dataset segments each 11-second recording into a 5-second history and a 6-second future.  The choice of polynomial degrees for agent trajectories has been extensively analyzed in our prior work \cite{yao_empirical_2023}, which examines the trade-off between representation fidelity and polynomial degree across various agent types and trajectory lengths in both the A2 and WO datasets. Based on the suggestions in \cite{yao_empirical_2023}, we represent different data types as follows:
\begin{itemize}[leftmargin=*]
    \item \textbf{Agent History}: Following the Akaike Information Criterion (AIC)  \cite{akaike_AIC_1973} from the study \cite{yao_empirical_2023}, we represent 5-second history trajectories
    of vehicles, cyclists, and pedestrians in A2 using optimal 5-degree polynomials. We also use the 5-degree polynomial for the ego vehicle. We track the
    control points of agent history with the method proposed in
    \cite{reichardt_trajectories_2022} and incorporate the observation noise models in \cite{yao_empirical_2023}.

    \item \textbf{Agent Future}: We model 6-second future trajectories as 6-degree polynomials -- one degree higher than suggested by AIC in \cite{yao_empirical_2023} to better capture complex trajectories. Bayesian Regression is applied to fit agent future trajectories, following priors and observation noise models from \cite{yao_empirical_2023}.
    
    \item \textbf{Map}: Map elements, such as lane segments and cross-
        walks, are represented with 3-degree polynomials, aligning with OpenDRIVE \cite{opendrive} standards. We fit the sample points of map elements via the total-least-squares method by Borges-Pastva \cite{borges_2002_total}.
\end{itemize}

\subsection{Model Architecture}
The pipeline of EP-Diffuser is illustrated in Figure \ref{fig: diffusion pipeline}. The model employs an encoder-denoiser architecture, with implementation details provided in Appendix \ref{sec: ep-diffuser implementation}. To mitigate variability in sampling procedures and accelerate the inference process, we adopt the Denoising Diffusion Implicit Models (DDIM) method \cite{song_denoising_2021} with 10 denoising steps -- consistent with the configuration of OptTrajDiff (details in Appendix \ref{sec: diffusion and denoising}).

\begin{figure}[thb]
\centering
\vspace{-0.5em}
\includegraphics[width=3.0in]{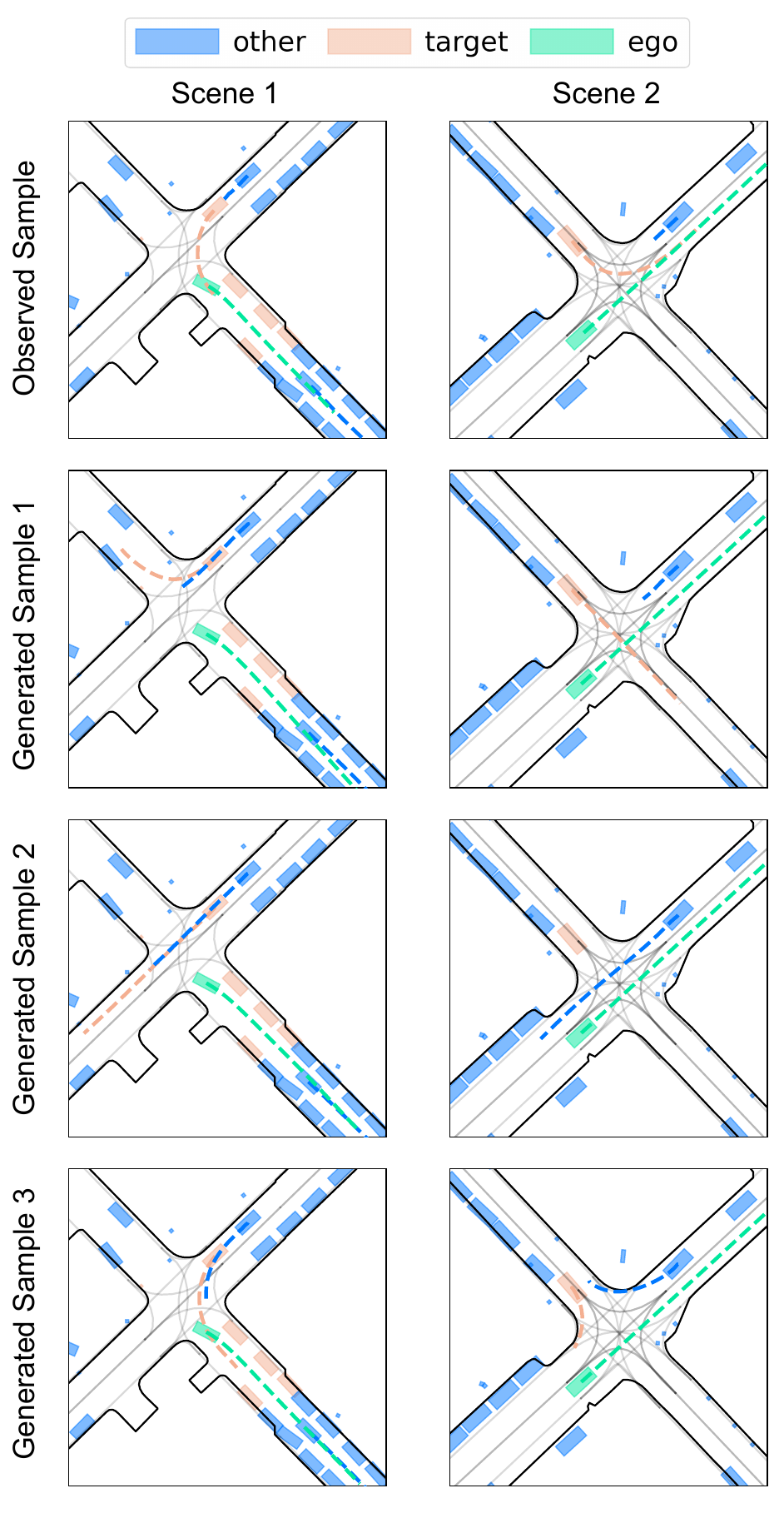}
\caption{Observed sample (ground truth) and generated traffic scene samples from EP-Diffuser for two traffic scenes in Argoverse 2, demonstrating EP-Diffuser's capability to generate diverse and plausible traffic scenes. \textbf{Dashed lines} represent the future trajectories of selected highly interactive agents.}
\label{fig: simulated traffic scenes}
\vspace{-0.5em}
\end{figure}

\section{Experimental Results}
\subsection{Experiment Setup}
\subsubsection{\textbf{In-Distribution Training and Testing on A2}}
Following the Argoverse 2 Motion Prediction Competition setup, models are required to output 6-second predictions given 5-second history. All models are trained with this setup from scratch with their original hyperparameters on the A2 training split containing 199{,}908 scenarios.

\begin{table*}[tb]
\vspace{+5pt}
\caption{Results of ``Sim Agents'' metrics with standard deviation over tested samples. The best value for each metric across models is highlighted in \textbf{bold}. \textbf{Upper Section}: Evaluated on the randomly selected $20\%$ subset from Argoverse 2 and Waymo validation sets. \textbf{Lower Section}: Evaluated on the 500 most challenging scenes from Argoverse 2 and Waymo validation sets. \textbf{In-Distribution Testing}: Performed on the Argoverse 2 validation set with a 6-second prediction horizon. \textbf{Out-of-Distribution Testing:} Performed on the homogenized Waymo validation set with a 4.1-second prediction horizon.}
\centering
    \begin{tabularx}{\textwidth}{c c c | *{6}{>{\centering\arraybackslash}X} c|c}
    \toprule
    \multicolumn{2}{c}{\multirow{2}{*}{train / test}} &\multirow{2}{*}{model} & realism & kinematic & interactive & map & minADE & coverage && \# model params \\
    &&& meta $\uparrow$ & metrics $\uparrow$ & metrics $\uparrow$ & metrics $\uparrow$ & [m] $\downarrow$& [m] && [million]\\
        \midrule
        \multirow{12}{*}{\rotatebox{90}{R20P}}&\multirow{6}{*}{A2 / A2} &FMAE-MA \cite{cheng_forecast_2023} &0.749$\scriptstyle\pm0.117$ & 0.459$\scriptstyle\pm0.181$ & 0.786$\scriptstyle\pm0.149$ & 0.867$\scriptstyle\pm0.187$ & 0.333$\scriptstyle\pm0.295$ & 2.000$\scriptstyle\pm3.672 $&&11.4 (6$\times$1.9)\\
        \cmidrule(lr){3-10} \cmidrule(lr){11-11}
        &&OptTrajDiff \cite{wang_optimizing_2025} & 0.795$\scriptstyle\pm0.091$ & 0.564$\scriptstyle\pm0.166$  & 0.805$\scriptstyle\pm0.117$ & \textbf{0.915}$\scriptstyle\pm0.121$ & \textbf{0.325}$\scriptstyle\pm0.277$ & 4.447$\scriptstyle\pm6.818$&&12.5 \\
        \cmidrule(lr){3-10} \cmidrule(lr){11-11}
        & &Seq-Diffuser (ablation) & 0.748$\scriptstyle\pm0.108$ & 0.414$\scriptstyle\pm0.195$ & 0.789$\scriptstyle\pm0.135$& 0.885$\scriptstyle\pm0.161$& 0.331$\scriptstyle\pm0.255$ & 1.718$\scriptstyle\pm3.714$ &&3.1\\
         \cmidrule(lr){3-10} \cmidrule(lr){11-11}
        && EP-Diffuser (ours) & \textbf{0.809}$\scriptstyle\pm0.087$ & \textbf{0.632}$\scriptstyle\pm0.149$ & \textbf{0.808}$\scriptstyle\pm0.105$& 0.913$\scriptstyle\pm0.129$& 0.398$\scriptstyle\pm0.340$ & 4.178$\scriptstyle\pm7.127$&&3.0\\
        \hhline{~==========}
        &\multirow{6}{*}{A2 / WO} & FMAE-MA \cite{cheng_forecast_2023} &0.709$\scriptstyle\pm0.102$  & 0.309$\scriptstyle\pm0.111$ & 0.800$\scriptstyle\pm0.125$ & 0.819$\scriptstyle\pm0.191$ & 0.363$\scriptstyle\pm0.203$ & 3.560$\scriptstyle\pm4.733$&&11.4 (6$\times$1.9)\\
        \cmidrule(lr){3-10} \cmidrule(lr){11-11}
        &&OptTrajDiff \cite{wang_optimizing_2025} & 0.768$\scriptstyle\pm0.068$ & 0.420$\scriptstyle\pm0.107$  & 0.827$\scriptstyle\pm0.087$ & 0.890$\scriptstyle\pm0.114$ & \textbf{0.310}$\scriptstyle\pm0.227$ & 3.496$\scriptstyle\pm4.692$&&12.5 \\
        \cmidrule(lr){3-10} \cmidrule(lr){11-11}
        &&Seq-Diffuser (ablation) &  0.688$\scriptstyle\pm0.079$ & 0.212$\scriptstyle\pm0.093$& 0.801$\scriptstyle\pm0.090$& 0.814$\scriptstyle\pm0.154$& 0.458$\scriptstyle\pm0.286$& 2.446$\scriptstyle\pm4.949$&&3.1\\
        \cmidrule(lr){3-10} \cmidrule(lr){11-11}
        &&EP-Diffuser (ours) & \textbf{0.788}$\scriptstyle\pm0.063$ & \textbf{0.491}$\scriptstyle\pm0.133$ & \textbf{0.834}$\scriptstyle\pm0.073$ & \textbf{0.900}$\scriptstyle\pm0.096$ & 0.348$\scriptstyle\pm0.307 $ & 3.224$\scriptstyle\pm5.603$&&3.0\\
        \bottomrule
        \multirow{12}{*}{\rotatebox{90}{C500}}&\multirow{6}{*}{A2 / A2}&FMAE-MA \cite{cheng_forecast_2023} &0.636$\scriptstyle\pm0.125$ & 0.320$\scriptstyle\pm0.118$ & 0.679$\scriptstyle\pm0.173$ & 0.760$\scriptstyle\pm0.241$ & 0.479$\scriptstyle\pm0.371$ & 2.780$\scriptstyle\pm4.155$&&11.4 (6$\times$1.9)\\
        \cmidrule(lr){3-10} \cmidrule(lr){11-11}
        &&OptTrajDiff \cite{wang_optimizing_2025} & 0.709$\scriptstyle\pm0.094$ & 0.459$\scriptstyle\pm0.119$  & \textbf{0.717}$\scriptstyle\pm0.128$ & \textbf{0.841}$\scriptstyle\pm0.176$ & \textbf{0.449}$\scriptstyle\pm0.329$ & 5.361$\scriptstyle\pm7.240$&&12.5 \\
        \cmidrule(lr){3-10} \cmidrule(lr){11-11}
        &&Seq-Diffuser (ablation) & 0.634$\scriptstyle\pm0.111$ & 0.279$\scriptstyle\pm0.111$ & 0.674$\scriptstyle\pm0.153$ & 0.786$\scriptstyle\pm0.212$ & 0.467$\scriptstyle\pm0.298$ & 2.020$\scriptstyle\pm4.188$  &&3.1\\
        \cmidrule(lr){3-10} \cmidrule(lr){11-11}
        &&EP-Diffuser (ours) & \textbf{0.713}$\scriptstyle\pm0.094$ & \textbf{0.507}$\scriptstyle\pm0.116$ & 0.707$\scriptstyle\pm0.124$& 0.838$\scriptstyle\pm0.178$& 0.546$\scriptstyle\pm0.315$ & 4.500$\scriptstyle\pm7.075$&&3.0\\
        \hhline{~==========}
         &\multirow{6}{*}{A2 / WO}& FMAE-MA \cite{cheng_forecast_2023} &0.630$\scriptstyle\pm0.123$ & 0.271$\scriptstyle\pm0.079$ & 0.716$\scriptstyle\pm0.159$ & 0.723$\scriptstyle\pm0.238$ & 0.426$\scriptstyle\pm0.229$ & 3.855$\scriptstyle\pm5.060$&&11.4 (6$\times$1.9)\\
        \cmidrule(lr){3-10} \cmidrule(lr){11-11}
        &&OptTrajDiff \cite{wang_optimizing_2025} & 0.721$\scriptstyle\pm0.085$ & 0.403$\scriptstyle\pm0.092$  & 0.771$\scriptstyle\pm0.106$ & 0.839$\scriptstyle\pm0.163$ & \textbf{0.357}$\scriptstyle\pm0.201$ & 3.189$\scriptstyle\pm4.527$&&12.5 \\
        \cmidrule(lr){3-10} \cmidrule(lr){11-11}
        &&Seq-Diffuser (ablation) & 0.630$\scriptstyle\pm0.098$ & 0.210$\scriptstyle\pm0.069$ & 0.739$\scriptstyle\pm0.126$ & 0.730$\scriptstyle\pm0.196$ & 0.448$\scriptstyle\pm0.206$ & 1.980$\scriptstyle\pm4.327$&&3.1\\
        \cmidrule(lr){3-10} \cmidrule(lr){11-11}
        &&EP-Diffuser (ours) & \textbf{0.742}$\scriptstyle\pm0.072$ & \textbf{0.456}$\scriptstyle\pm0.099$ & \textbf{0.782}$\scriptstyle\pm0.090$ & \textbf{0.854}$\scriptstyle\pm0.134$ & 0.372$\scriptstyle\pm0.175$ & 2.786$\scriptstyle\pm5.244$ &&3.0\\
        \bottomrule
    \end{tabularx}
    \label{tab: sim agent 20 per}
    \vspace{-1.5em}
\end{table*}

The ``Sim Agents'' evaluation requires 32 modeled traffic scene samples for metric computation. For generative models, we randomly sample 32 generated traffic scenes. For FMAE-MA, which by design outputs 6 different predictions, we use an ensemble of 6 independently trained models, each initialized with a different seed, predicting 36 traffic scenes in total. From these, we select the 32 predictions with the highest predicted probabilities for evaluation.

\subsubsection{\textbf{Out-of-Distribution Testing on WO}}
Cross-dataset testing presents challenges due to inconsistencies in data formats and prediction tasks. To address this, we adopt the homogenization protocol from \cite{yao_improving_2024} to enable cross-dataset evaluation between the A2 and WO datasets. This protocol aligns WO's prediction task with the A2's competition setup by incorporating a 5-second history. Since WO recordings are shorter (9.1 seconds), we evaluate only the first 4.1 seconds of the 6-second predictions. Additionally, models are restricted to only considering lane centers and crosswalks as available map information due to their availability across both datasets. 

For OoD testing, we apply the same sampling strategies as in A2. All three models are trained on the homogenized A2 training split and tested on OoD samples from the homogenized WO validation split. Results are reported based on the 4.1-second prediction.

\subsubsection{\textbf{Test Data Selection}}
Due to the computational cost of ``Sim Agents'' metric calculations, we subsample the validation data from both the A2 and WO datasets under two distinct settings:
\begin{itemize}[leftmargin=*]
    \item \textbf{20\% Random Subsample:} We randomly select $20\%$ of the validation set from each dataset, corresponding to 5{,}000 samples from A2 and 8{,}400 samples from WO. This setting aims to reflect the model's overall performance across typical scenes presented by datasets. We refer to this subset as \textbf{R20P} in the remainder of the paper.

    \item \textbf{500 Most Challenging Scenes:} We identify and select the 500 most difficult traffic scenes in each validation set based on the largest deviations between ground truth trajectories and those generated by a constant velocity model, as measured by the realism meta metric. This setting emphasizes model performance under more complex and demanding conditions. We refer to this subset as \textbf{C500} in the remainder of the paper.
\end{itemize}

\subsubsection{\textbf{Coverage Metric}}
Inspired by the inter-policy diversity metric in \cite{shiroshita_behaviorally_2020}, we compute the average pairwise distance between the final positions of the 32 sampled trajectories to measure the trajectory diversity. This metric serves as an \textbf{auxiliary} indicator and is not used as a standalone measure of performance -- since high coverage is only meaningful when accompanied by high plausibility.

\subsection{Comparison with SotA Models on A2}
Table \ref{tab: sim agent 20 per} (\textbf{``A2/A2'' sections}) summarizes the in-distribution results of all models tested on the A2 validation set across both the R20P and C500 subsets. Figure \ref{fig: simulated traffic scenes} visualizes multiple samples generated by EP-Diffuser.

We observe a reversed ranking when evaluating prediction plausibility and $minADE$. For example, on R20P, despite its smaller model size ($3$ million parameters) and the largest $minADE$ (\SI{0.398}{\meter}), EP-Diffuser achieves the highest realism meta score at $0.809$ and performs comparably to OptTrajDiff in terms of agent interaction and map adherence, while also maintaining high coverage (\SI{4.178}{\meter}). Notably, EP-Diffuser demonstrates superior agent kinematics ($0.632$), outperforming OptTrajDiff ($0.564$) and FMAE-MA ($0.456$).

\begin{figure}[thb]
\vspace{+2pt}
\centering
\includegraphics[width=3.2in]{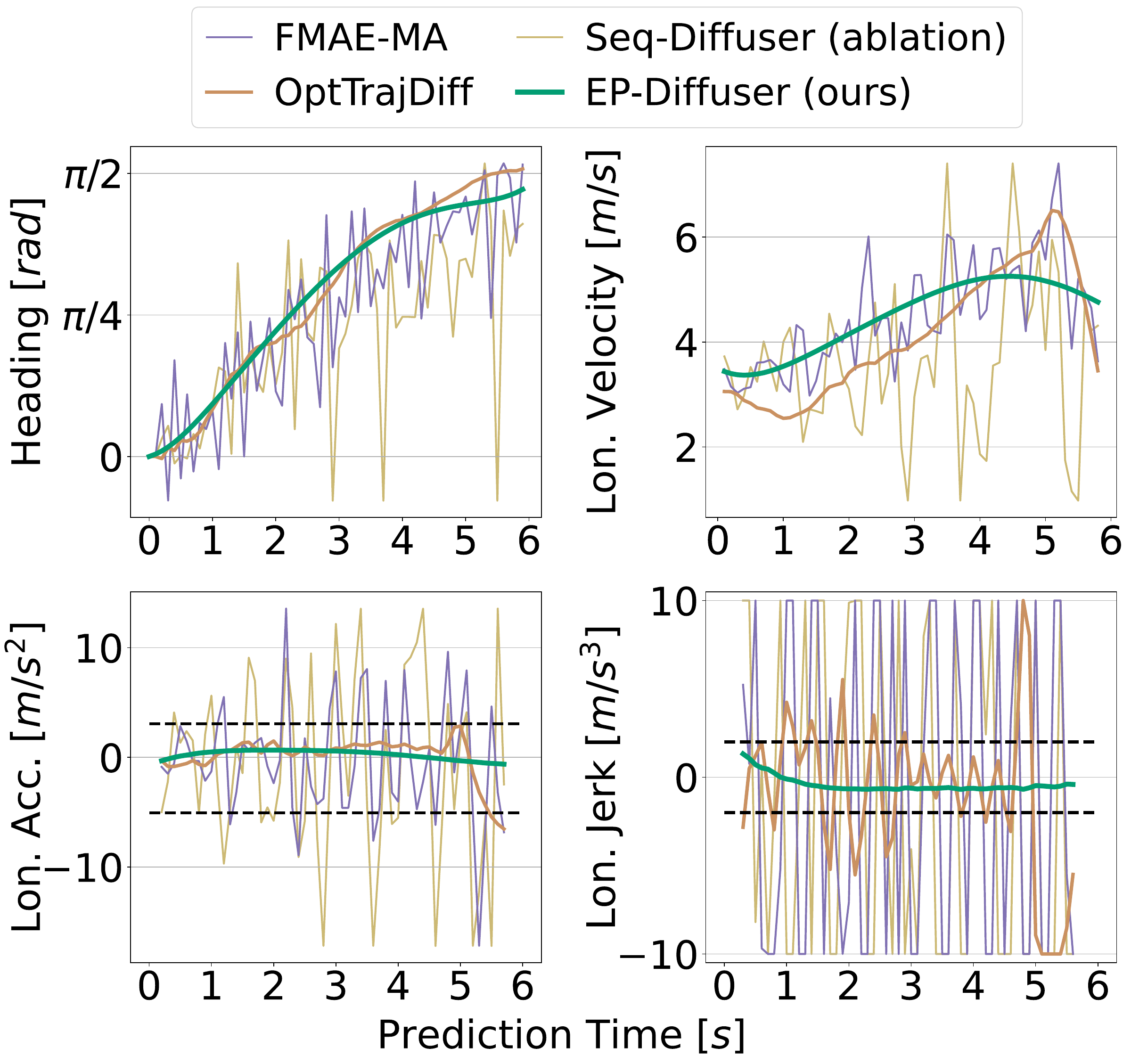}
\caption{Predicted heading, longitudinal velocity, acceleration, and jerk for a vehicle's left turn maneuver over a 6-second time horizon in A2. For clarity, the results of Seq-Diffuser, as well as the jerk outputs of FMAE-MA and OptTrafDiff, are clipped. \textbf{Dashed lines} indicate the ranges of longitudinal acceleration and jerk for aggressive human drivers based on \cite{bae_self_2020}, suggesting that the agent kinematics of EP-Diffuser are the most plausible.}
\label{fig: simulated agent heading}
\vspace{-1.0em}
\end{figure}

As an illustrative example, Figure \ref{fig: simulated agent heading} visualizes the vehicle kinematics during a left-turn maneuver in A2. EP-Diffuser produces the most plausible agent kinematics according to the measures in \cite{bae_self_2020}.

\begin{figure}[th]
\vspace{+2pt}
\centering
\includegraphics[width=3.0in]{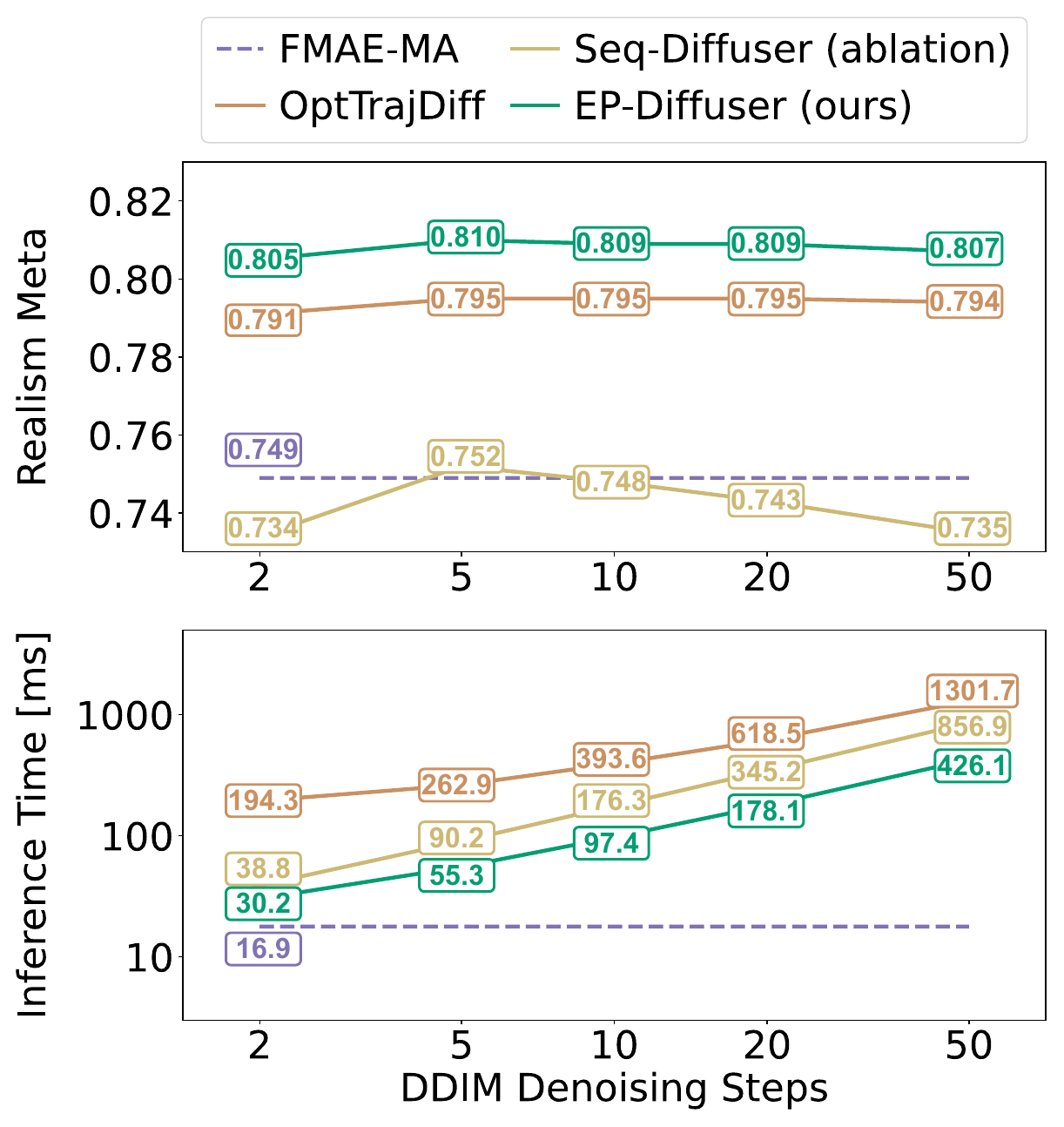}
\caption{Realism meta score on A2 R20P and inference time for diffusion models across different DDIM denoising steps, with FMAE-MA as the regression-based baseline
for comparison. Both the denoising steps and inference time are shown on \emph{log scale}. Inference time is measured by predicting 6 samples of an Argoverse 2 traffic scene with 50 agents and 150 map elements, using a single A10G GPU.}
\label{fig: influence of DDIM steps}
\vspace{-1.0em}
\end{figure}

Figure \ref{fig: influence of DDIM steps} reports the realism meta scores and inference time of diffusion-based models over a range of DDIM denoising steps, with FMAE-MA as the regression-based baseline for comparison. Across all denoising step configurations, EP-Diffuser consistently outperforms OptTrajDiff in both metrics, achieving higher realism scores with lower inference time. While EP-Diffuser's inference time exceeds FMAE-MA's, it reaches peak performance with 5 denoising steps in just \SI{55.3}{\milli\second}, making it viable for real-time applications.

\subsection{OoD Testing on WO}
In the OoD Setting, the model trained on A2 is asked to generate traffic scene continuations for independent scenes taken from the WO dataset -- thus removing any sort of shared bias between training and test data. The OoD testing results are presented in the \textbf{``A2/WO'' sections} in Table \ref{tab: sim agent 20 per}.

In the OoD setting, EP-Diffuser maintains the top scores in realism meta and agent kinematics. Additionally, it also demonstrates improved performance by achieving the best agent interactions and map adherence scores on both subsets. Furthermore, EP-Diffuser outperforms FMAE-MA in $minADE$ and closely matches OptTrajDiff (\SI{0.372}{\meter} vs. \SI{0.357}{\meter} on C500, respectively)   with high coverage (\SI{2.786}{\meter}). These results demonstrate EP-Diffuser's ability to learn robust, transferable representations from training data, highlighting its enhanced generalization beyond dataset-specific patterns.

\subsection{Ablation Study}
To evaluate the impact of polynomial representations, we introduce a variant, \textbf{Seq-Diffuser}, which replaces polynomial inputs and outputs in EP-Diffuser with sequence-based representations. Only minor modifications to the model architecture are required to accommodate the increased dimensionality of the sequence-based representation. The training and diffusion settings are identical for both EP-Diffuser and Seq-Diffuser. Comparative results are presented in Table \ref{tab: sim agent 20 per}.

While Seq-Diffuser achieves better in-distribution accuracy with lower $minADE$ (e.g. \SI{0.331}{m} vs. \SI{0.398}{m} on R20P), it exhibits a notable degradation in both the plausibility and diversity of the generated traffic scenes across both subsets. Moreover, Seq-Diffuser generalizes poorly, as reflected by its lower performance in plausibility and higher OoD $minADE$ (e.g. \SI{0.458}{m} vs. \SI{0.348}{m} on R20P). It also incurs a higher inference time, as shown in Figure \ref{fig: influence of DDIM steps}.

These results indicate that, although sequence-based representations may capture finer-grained details beneficial for pointwise accuracy, polynomial representations lead to better plausibility, diversity, and OoD generalization in multi-agent scene generation.


\section{Conclusion}    
Traffic scene prediction constitutes a fundamental challenge in autonomous driving due to its inherently multi-modal nature. Most motion prediction competitions aim for prediction accuracy, encouraging models to focus on reproducing observed behaviors rather than capturing the diversity of plausible future scene evolutions. In this work, we introduced EP-Diffuser, a novel diffusion-based framework that leverages polynomial representations to efficiently model agent trajectories and road geometry. Through extensive experiments on Argoverse 2 and Waymo Open datasets, we demonstrated that EP-Diffuser not only improves the plausibility of generated traffic scenes but also generalizes well to out-of-distribution (OoD) environments. Notably, EP-Diffuser's computational efficiency addresses a fundamental limitation of diffusion models, making it viable for real-time applications.

Future work will focus on 
refining evaluation methodologies to further bridge the gap between prediction accuracy and coverage of plausible alternatives for real-world feasibility. Additionally, expanding EP-Diffuser to incorporate uncertainty-aware decision-making for autonomous vehicles presents an exciting direction for enhancing robustness in motion planning.


\section*{ACKNOWLEDGMENT}
The authors would like to thank Jonas Neuhoefer and Andreas Philipp for many helpful discussions. We also gratefully acknowledge the Gauss Center for Supercomputing for providing compute on JUWELS at JSC.  This work is funded by the German Federal Ministry for Economic Affairs and Energy within the project ``NXT GEN AI METHODS''.

\bibliographystyle{IEEEtran}
\bibliography{references}

\begin{appendices}
\section{Implementation Details}
\label{sec: ep-diffuser implementation}

\subsection{Diffusion and Denoising}
\label{sec: diffusion and denoising}
We perform diffusion-denoising on the polynomial parameters (control points) that represent each agent’s future trajectory. Specifically, we operate on the displacement vectors between consecutive control points of 6-degree Bernstein polynomials representing 2D future trajectories, denoted as $\boldsymbol{\delta}^{fut} \in \mathbb{R}^{12}$ for each agent.

Following the practice of Denoising Diffusion Probabilistic Models (DDPM) \cite{ho_denoising_2020}, we denote the diffused $\boldsymbol{\delta}^{fut}$ for the $i$-th agent at $s$-th diffusion step as $\boldsymbol{\delta}^{fut}_{s,i} \in \mathbb{R}^{12}$. Here, $
s=0$ corresponds to the fitted polynomial parameters without added noise. We can write $\boldsymbol{\delta}^{fut}_{s,i}$ as a linear combination of the initial parameters $\boldsymbol{\delta}^{fut}_{0,i}$ and a Gaussian noise $\boldsymbol{\epsilon}_i$:
\begin{equation}
\begin{aligned}
\label{eqn: diffusion linear}
\boldsymbol{\delta}^{fut}_{s,i} &= \sqrt{\Bar{\alpha}_s}\boldsymbol{\delta}^{fut}_{0,i} + \sqrt{1-\Bar{\alpha}_s}\boldsymbol{\epsilon}_i, \quad \text{where } \boldsymbol{\epsilon}_i \sim \mathcal{N}(\mathbf{0}, \mathbf{I}). \\
\end{aligned}
\end{equation}
\noindent where $\Bar{\alpha}_{s}$ is the noise-scheduling parameter at diffusion step $s$ and controls the diffusion process.

We apply a total of $S=1000$ diffusion steps to gradually transition from the data distribution $q(\boldsymbol{\delta}^{fut}_{0,i})$ to the target prior distribution $\mathcal{N}(\mathbf{0}, \mathbf{I})$. For the backward denoising process, we adopt Denoising Diffusion Implicit Models (DDIM) \cite{song_denoising_2021} -- the same procdedure as OptTrajDiff -- with 10 denosing steps. The future trajectories of agents are iteratively denoised by predicting the added noise $\hat{\boldsymbol{\epsilon}}_i$ for each agent and subtracting $\hat{\boldsymbol{\epsilon}}_i$ from $\boldsymbol{\delta}^{fut}_{s,i}$ at each step.

\subsection{Encoder}
We employ the encoder architecture of our EP model \cite{yao_improving_2024} and adopt the ``query-centric'' design of QCNet \cite{zhou_query_2023}, adapting it to operate on polynomial representations. 

For all agents $\mathcal{A}$ in a scene, we encode the 2D displacement vectors of agent history control points, modeled as 5-degree polynomials, denoted as $\boldsymbol{\Delta}^{hist} \in \mathbb{R}^{A \times 10}$, where $A$ is the number of agents. Additionally, we encode the time window of each agent’s appearance in history, represented as $\boldsymbol{TW} \in \mathbb{R}^{A \times 2}$, along with the agent category information. These features are summarized to form the agent condition tokens $\boldsymbol{C}^{agent} \in \mathbb{R}^{A \times D}$, where $D$ is the hidden dimension. 

Similarly, for all map elements $\mathcal{M}$ in a scene, we encode their control point vectors of 3-degree polynomials, denoted as $\boldsymbol{\Delta}^{map} \in \mathbb{R}^{M \times 6}$, along with the corresponding map element categories. These features are summarized to form the map condition tokens $\boldsymbol{C}^{map} \in \mathbb{R}^{M \times D}$, where $M$ is the number of map elements. 

\subsection{Denoiser}
\begin{figure}[thb]
\centering
\includegraphics[width=0.48\textwidth]{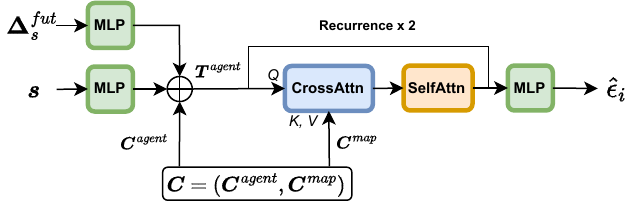}
\caption{Denoiser architecture of EP-Diffuser.}
\label{fig: denosier}
\vspace{-1.0em}
\end{figure}

We visualize the denoiser architecture in Figure \ref{fig: denosier}. The denoiser processes the vectors between control points of noised 6-degree future trajectories, denoted as $\boldsymbol{\Delta}^{fut}_s \in \mathbb{R}^{A \times 12}$, along with the diffusion step indices for each agent $\boldsymbol{s} \in \mathbb{R}^{A}$. These inputs are embedded and summarized with the agent condition tokens $\boldsymbol{C}^{agent}$ to form the agent tokens $\boldsymbol{T}^{agent} \in \mathbb{R}^{A \times D}$. Multiple attention blocks based on Transformer \cite{vaswani_attention_2017} perform the agent-map and agent-agent attentions sequentially to update $\boldsymbol{T}^{agent}$. Finally, the updated agent tokens $\boldsymbol{T}^{agent}$ is decoded to predict the added noise $\hat{\boldsymbol{\epsilon}}_i$ for each agent.

\subsection{Training Loss}
Following DDPM \cite{ho_denoising_2020}, the EP-Diffuser is trained to minimize the mean squared error (MSE) between the added noise $\boldsymbol{\epsilon}_i$ and predicted noise $\hat{\boldsymbol{\epsilon}}_i$ averaged across all agents: $\mathcal{L} = \frac{1}{A}\sum_{i=1}^A || \boldsymbol{\epsilon}_i - \hat{\boldsymbol{\epsilon}}_i ||^2_2$.

\subsection{Training Setup}
We report the training setup for EP-Diffuser in Table \ref{tab: setting for EP-Diffuser}. The noise scheduling parameter is expressed as $\bar{\alpha}_s = \Pi^s_k \alpha_k$, where $\alpha_s=1-\beta_s$. 

\begin{table}[ht]
\caption{EP-Diffuser Training Setup}
\vspace{-0.0em}
\centering
\begin{tabularx}{3.4in}{c  >{\centering\arraybackslash}X }
\Xhline{3\arrayrulewidth}
hidden dimension $D$ &  128\\
\hline
\multirow{2}{*}{$\beta_s$} &  $ s*(\beta_{end}-\beta_{start})/S + \beta_{start} , \quad \text{with}$\\
&  $\beta_{end}=0.2, \beta_{start}=1e-5, S=1000$ \\
\hline
optimizer &  AdamW\\
\hline
learning rate & 5e-4\\
\hline
learning rate schedule & cosine\\
\hline
batch size &  32\\
\hline
training / warmup epochs  & 64 / 10\\
\hline
dropout &  0.1 \\
\Xhline{3\arrayrulewidth}
\vspace{-1.5em}
\end{tabularx}
\label{tab: setting for EP-Diffuser}
\end{table}

\subsection{Post-processing}
We observe that stationary agents in recorded ground truth often exhibit minor positional shifts and unrealistic rotations, which can lead to unnatural behaviors in predicted scenes across all models. To address this, we adopt a lightweight post-processing step inspired by prior work \cite{zeng_dsdnet_2020}:
\begin{itemize}[leftmargin=*]
    \item \textbf{Stationary Agent Correction}: For each predicted trajectory, if an agent moves less than \SI{1}{\meter} over the prediction horizon, we classify it as non-moving and retain its last measured position and heading.
\end{itemize}
\noindent This ensures physically consistent behavior for stationary agents without modifying the model’s core predictions. For fairness, we apply this post-processing step uniformly across both our model and benchmark models.
\end{appendices}

\end{document}